%% file: cvpr.tex
\documentclass[final]{cvpr}

\makeatletter
\renewcommand{\paragraph}{%
  \@startsection{paragraph}{4}%
  {\z@}{1ex \@plus 1ex \@minus .2ex}{-1em}%
  {\normalfont\normalsize\bfseries}%
}
\makeatother
\usepackage{times}
\usepackage{epsfig}
\usepackage{graphicx}
\usepackage{amsmath}
\usepackage{amssymb}
\usepackage{subfigure}
\usepackage{multirow}
\usepackage{booktabs}
\usepackage{mwe}
\usepackage{subfigure}
\usepackage{paralist}
\usepackage{algorithm}
\usepackage{algorithmic}

\usepackage[pagebackref=true,breaklinks=true,colorlinks,bookmarks=false]{hyperref}
\usepackage[capitalise]{cleveref}

\def\cvprPaperID{1901} 
\def\confYear{CVPR 2021}

\begin{document}

\title{VersatileGait: A Large-Scale Synthetic Gait Dataset with Fine-Grained Attributes and Complicated Scenarios}


\author{Huanzhang Dou\textsuperscript{1}, Wenhu Zhang\textsuperscript{1}, Pengyi Zhang\textsuperscript{1}, Yuhan Zhao\textsuperscript{1},
\\Songyuan Li\textsuperscript{1}, Zequn Qin\textsuperscript{1}, Fei Wu\textsuperscript{1,3}, Lin Dong\textsuperscript{2}, Xi Li\textsuperscript{1,3}\thanks{Corresponding author. E-mail: xilizju@zju.edu.cn.}
\\
\textsuperscript{1}Zhejiang University, \textsuperscript{2}Merit Interactive Co., Ltd.\\
\textsuperscript{3}Shanghai Institute for Advanced Study, Zhejiang University,\\

}

\maketitle

\begin{abstract}
With the motivation of practical gait recognition applications, we propose to automatically create a large-scale synthetic gait dataset (called VersatileGait) by a game engine, which consists of around one million silhouette sequences of 11,000 subjects with fine-grained attributes in various complicated scenarios. Compared with existing real gait datasets with limited samples and simple scenarios, the proposed VersatileGait dataset possesses several nice properties, including huge dataset size, high sample diversity, high-quality annotations, multi-pitch angles, small domain gap with the real one, etc. Furthermore, we investigate the effectiveness of our dataset (e.g., domain transfer after pretraining). Then, we use the fine-grained attributes from VersatileGait to promote gait recognition in both accuracy and speed, and meanwhile justify the gait recognition performance under multi-pitch angle settings. Additionally, we explore a variety of potential applications for research. Extensive experiments demonstrate the value and effectiveness of the proposed VersatileGait in gait recognition along with its associated applications. We will release both VersatileGait and its corresponding data generation toolkit for further studies.

   \end{abstract}

\input{data/intro}

\input{data/related}
\input{data/data}

\input{data/effective}
\input{data/applications}

\input{data/future}

\input{data/conclusion}



{\small
\bibliographystyle{ieee_fullname}
\bibliography{egbib}
}

\end{document}

%% file: data/intro.tex
\section{Introduction}


\begin{figure}[htbp]
  \centering
  \subfigure[Real dataset collection.]{
  \includegraphics[width=7.5cm]{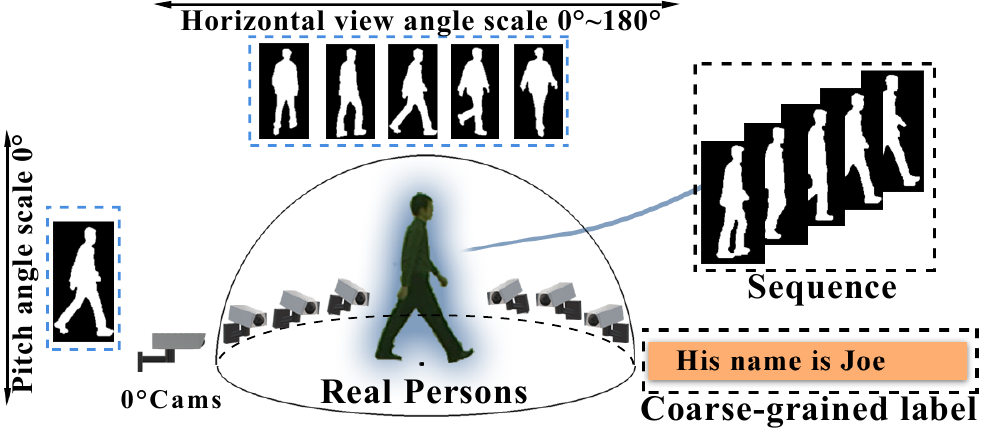}
  \label{realdata}
  }
  \quad
  \subfigure[Synthetic data generation.]{
  \includegraphics[width=7.5cm]{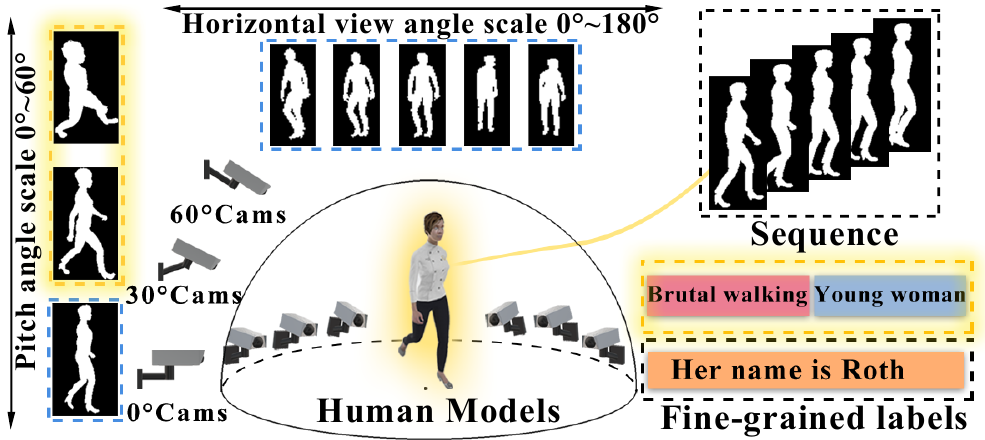}
  \label{fakedata}
  }
 \caption{The comparison of real and synthetic datasets generation. VersatileGait differs in the fine-grained attributes and complicated scenarios, which are highlighted with yellow color.} 
  \end{figure}
 
  As an important and challenging problem in computer vision, gait recognition~\cite{chao2019gaitset,connor2018biometric,Fan_2020_CVPR,gafurov2007survey,pushparani2012survey,8600947,sugandhi2016feature,wan2018survey} aims to identify the individual walking pattern, and 
  has a wide range of applications such as visual surveillance~\cite{bouchrika2018survey}, security checks~\cite{6883181}, and video retrieval~\cite{samangooei2010performing}. Compared with other biometric recognition approaches, it has the following advantages: 1) non-explicit cooperation with humans; 2) long-distance perception; 3) robustness to changes in accessories.

  In principle, a gait pattern in the wild is usually represented as a sequence of human silhouettes (\ie, binary masks without textures), which vary greatly with respect to many complicated intrinsic and extrinsic factors. Specifically, intrinsic factors usually refer to individual-specific attributes (\eg, genders, walking styles, and complex accessories), which affect the silhouettes intrinsically due to the differences in muscle action and body appearance. Extrinsic factors often correspond to camera settings (\eg, horizontal views and pitch angles), which may lead to silhouette distortion. Therefore, effective gait recognition in the wild requires datasets to cover a wide variety of fine-grained attributes and complicated scenarios.

  However, existing datasets are confronted with the following three limitations: 
  \begin{inparaenum}[1)]
   \item simple annotations with only ID labels (without individual attributes); 
   \item ideal scenarios with a single camera pitch angle, which are shown in~\cref{realdata}; 
   \item small scale of data in terms of either the individual ID number or the silhouette sequence number for each ID.
  \end{inparaenum}
  For example, the CASIA-B dataset is composed of 124 individual IDs, 13632 silhouette sequences with clothing and bag condition. In the OU-MVLP dataset, each subject only walks twice without the change of clothing or bag. As a result, these datasets are inadequate for practical gait recognition.  There is an urgent need for building a large-scale gait dataset with fine-grained attributes and complicated scenarios.

  In practice, constructing a real dataset for gait recognition is difficult due to the following factors:
  1) data collection is very expensive and time-consuming, \eg, collecting OU-MVLP takes over 11 months with a huge cost~\cite{takemura2018multi}; 2) privacy issues for individual IDs; 3) limited data processing, which causes silhouettes quality damages, \eg, irregular cavity and body missing. Thus, we turn to build the dataset by computer simulation, which is a cheap and convenient method.
  
   We present a large-scale synthetic dataset called VersatileGait based on virtual data simulation with a game engine,  as shown in \cref{fakedata}. Specifically, to follow a high-quality realistic standard, our data simulation process is composed of the following steps: 3D model generation, walking animation collection, animation retargeting, complicated scene simulation, and silhouettes capture. Such a synthetic dataset distinguishes in:
   \begin{inparaenum}[1)]
     \item fine-grained annotations thanks to parameterized human models;
     \item complicated scenarios 
      for different demands in terms of customized scenes which are seamless to real applications;
     \item high quality, low costs, and no privacy issues;
     \item small domain gap. Thanks to the colorless and textureless properties, the silhouette sequences generated by virtual data simulation are close to the real ones. 
     This phenomenon makes it more suitable to generate synthetic data on gait recognition than the tasks which utilize RGB datasets.
    \end{inparaenum}

  Based on VersatileGait, we improve the performance of the existing models by a considerable margin and propose numerous practical applications. Firstly, VersatileGait could be used to pretrain a deep learning model, which is then finetuned by task-specific real datasets. In this case, 
  the performance of the SOTA method~\cite{Fan_2020_CVPR} increases by 1.1\% of rank-1 accuracy on CASIA-B. Besides, we could set up a new attribute classification task together with the primary gait recognition, resulting in a multi-task learning problem.
  The results of attribute prediction could be used for gait retrieval acceleration. 
  Meanwhile, we conduct the performance evaluations of gait recognition in complicated scenarios (\eg, multi-pitch angles). 
  Last but not least, we provide some potential applications, \ie, the multi-person gait recognition and individual-related attributes for disentanglement learning.
   Therefore, VersatileGait is more suitable for real scenarios, and also promotes the gait recognition literature from the research perspective. 
   We will also release our data generation code and datasets to the research community.

 On the whole, the main contributions of this work are summarized as follows:
  \begin{itemize}
      \item We propose to develop a large-scale synthetic gait dataset called VersatileGait, which covers fine-grained attributes and
      complicated scenarios. To our knowledge, VersatileGait is the first synthetic gait dataset for research use and practical applications.
      \item We enrich some new applications for potential research directions and also evaluate the performances of different gait recognition methods in complicated scenarios.
      \item We conduct extensive experiments to demonstrate the effectiveness of using VersatileGait, which improves the rank-1 accuracy of mainstream methods by a considerable margin.
  \end{itemize}

%% file: data/related.tex
\section{Related Work}
\subsection{Gait Recognition}
Gait recognition is a type of biometric authentication that identifies people by their walking patterns. Most previous works can be divided into two categories.

\paragraph{Model-Based Methods} This paradigm~\cite{6117582, bodor2009view,4378964, 1613073,liao2020model} concentrates on fitting articulated human body to images and uses 2D body joint features. They are robust to the covariates such as bags and clothes. However, this fashion relies heavily on the high-resolution images and the accurate pose estimation results

\paragraph{Appearance-Based Methods} These methods~\cite{chao2019gaitset,Fan_2020_CVPR,5299188,6680737,6478807,Li_2020_CVPR,10.1007/11744078_12,gait-recognition-via-disentangled-representation-learning} use the silhouette sequence, gait energy image (GEI)~\cite{1561189}, or gait entropy image (GEnI)~\cite{5522296} as inputs and accomplish gait recognition in an end-to-end fashion. These methods are popular due to their flexibility, effectiveness, and conciseness.

\subsection{Real Datasets in Gait Recognition}
Gait data in the real-world is usually collected in the neat laboratory followed by data processing, \eg, background subtraction. We introduce two mainstream datasets and list the statistics results of most of the gait datasets in \cref{datasetcompare}.

\paragraph{CASIA-B~\cite{1699873}} This dataset is the most frequently used one, which contains 124 subjects, 3 walking conditions (NM, BG, CL), and 11 horizontal views. Each subject contains 6 normal (NM) sequences, 2 walking with bag (BG) sequences, and 2 wearing clothing (CL) sequences.

\paragraph{OU-MVLP~\cite{takemura2018multi}} It contains 10307 subjects, and each subject owns 14 horizontal views and 2 sequences per view. OU-MVLP is the largest public gait dataset. 

\begin{figure}[htbp]
      \centering
      \subfigure[Irregular Cavity]{
      \includegraphics[width=2.1cm]{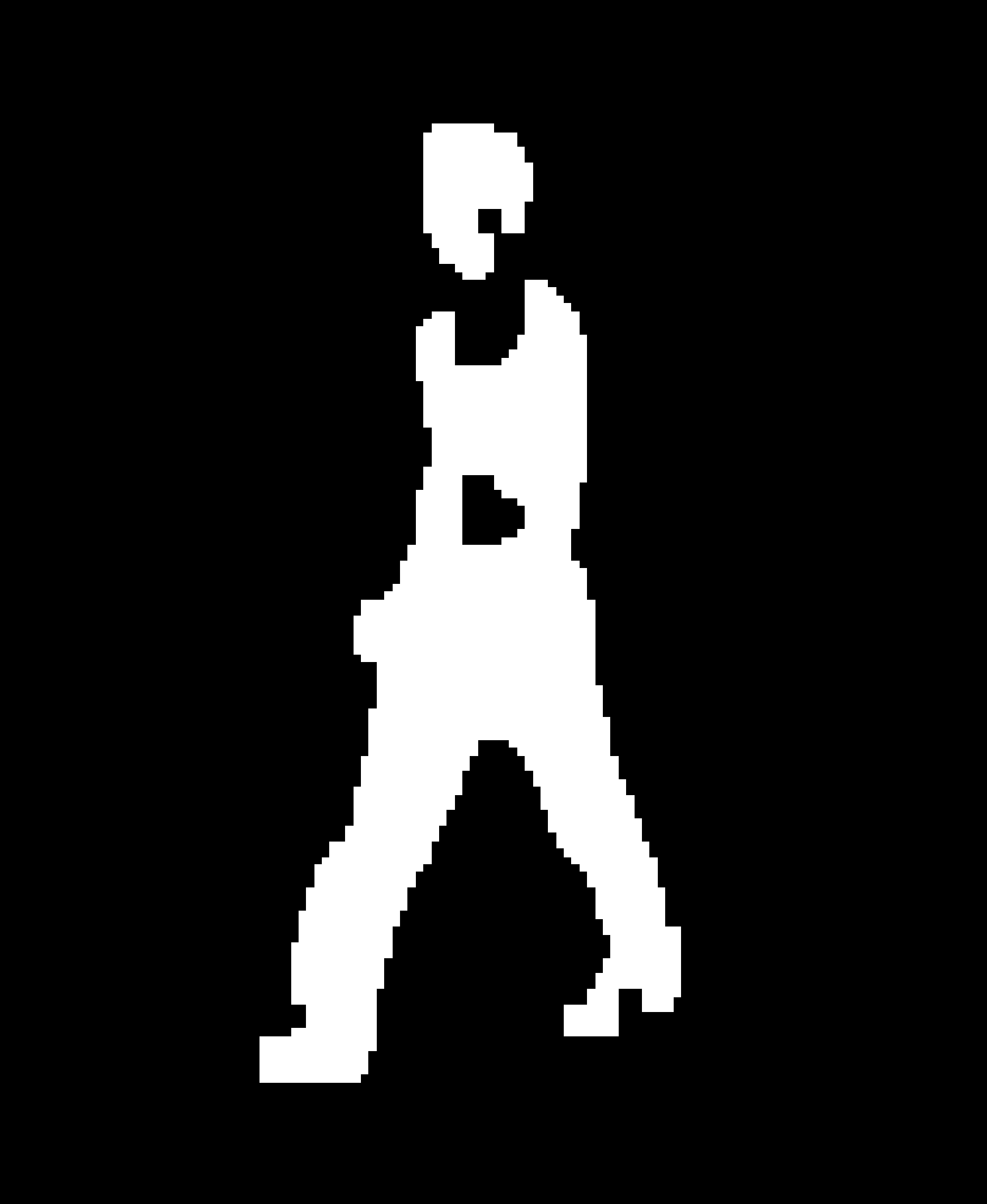}
      }
      \quad
      \subfigure[Body Missing]{
      \includegraphics[width=2.1cm]{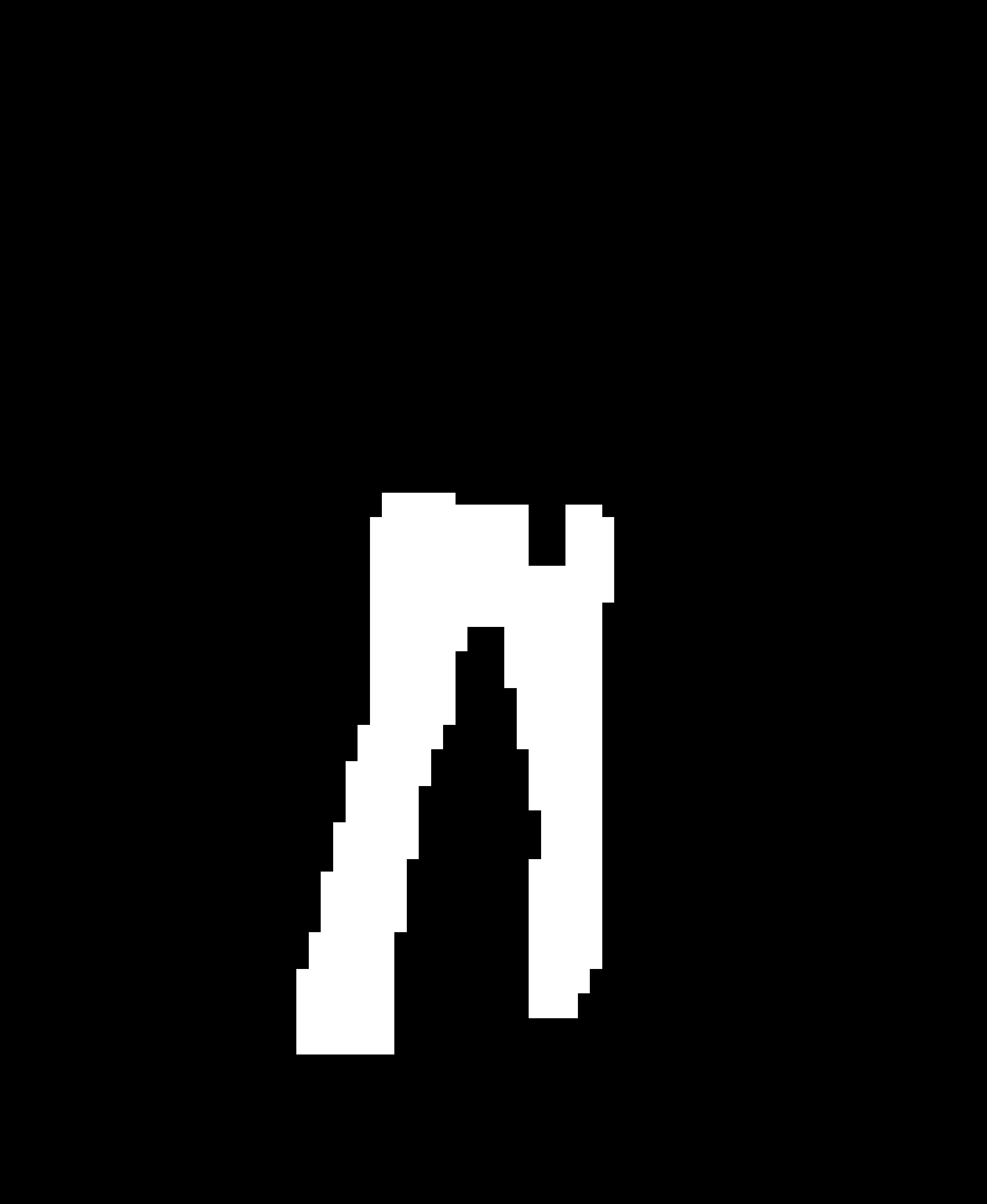}
      }
      \quad
      \subfigure[Clothing Issue]{
      \includegraphics[width=2.1cm]{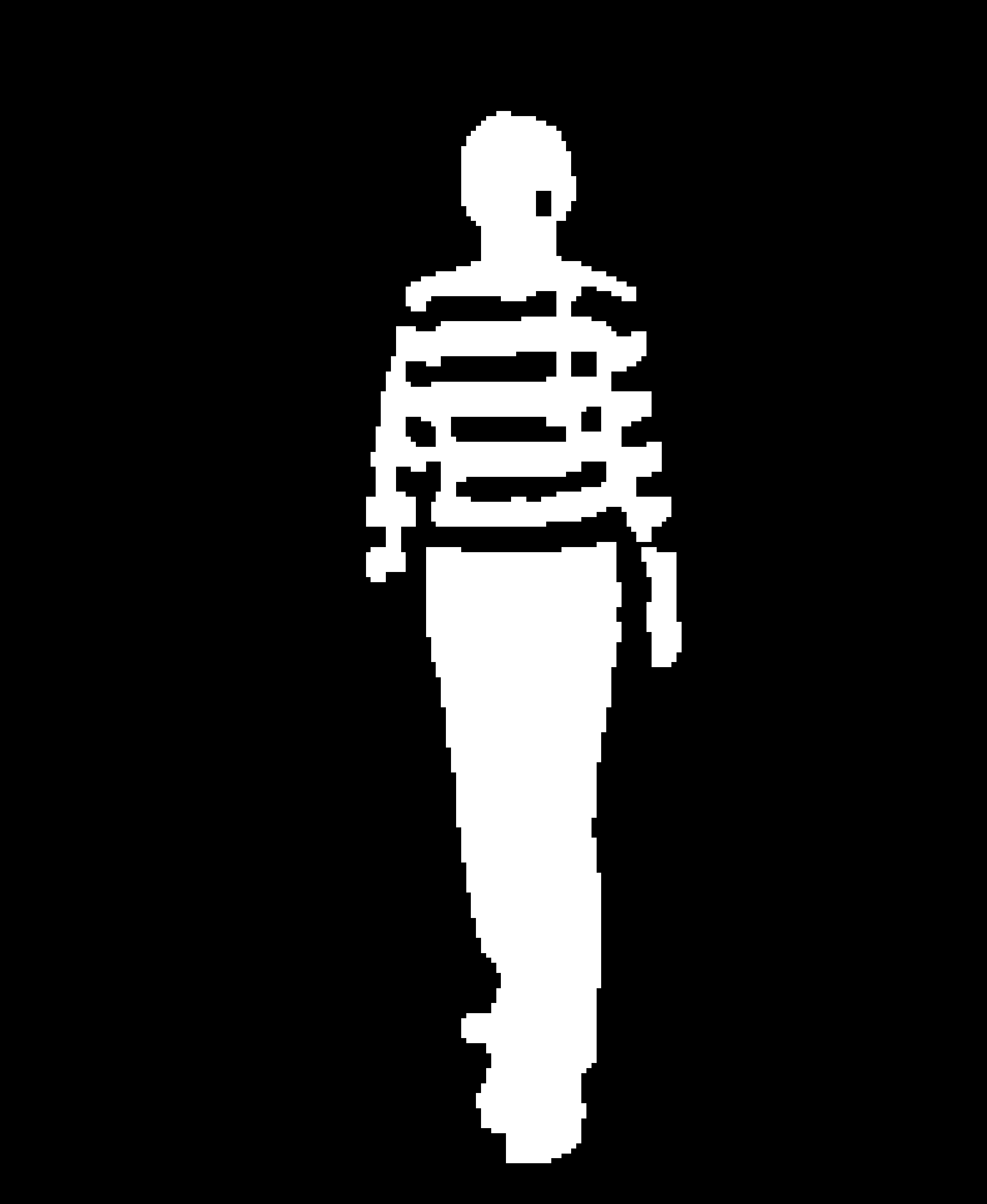}
      }
      \quad
      \caption{Common image quality problems in real dataset.}
      \label{real}
      \end{figure}

Previous datasets are collected under fairly simple scenarios and image quality problems exist due to data processing, which is shown in \cref{real}.

\begin{table}[t]
   \centering
   \caption{The comparison of reviewed databases in gait recognition. The VP (viewpoints) of cameras include the horizontal and vertical angles. VersatileGait distinguishes in ID, SQ (sequences), VP (viewpoints), BG/CL (bag, clothing), and contains rich features such as AT (attributes), PA (multi-pitch angles), MP (multi-person gait). The minimum number of ID and sequences are 20 and 240.}
   \setlength{\tabcolsep}{0.8mm}
   \begin{tabular}{lccccccc}
   \toprule
   Dataset                 & ID      & SQ       & VP     & BG/CL      & AT & PA & MP \\ 
   \midrule
   CASIA-A~\cite{wang2002gait}                & 1$\times$          & 1$\times$           & 3           &     &             &             &             \\ 
   CMU MoBo~\cite{Gross-2001-8255}                & 1$\times$          & 3$\times$             & 6           &           &             &             &             \\ 
   USF~\cite{1374864}                     & 6$\times$         & 8$\times$              & 2           &  $\checkmark$          &             &             &             \\ 
   CASIA-B~\cite{1699873}                & 6$\times$         & 57$\times$             & 11          & $\checkmark$         &             &             &             \\ 
   WOSG~\cite{decann2013investigating}                    & 8$\times$         & 3$\times$             & 8           &           &             &             &             \\ 
 
   CASIA-C~\cite{1699695}                & 8$\times$         & 6$\times$             & 1           & $\checkmark$         &             &             &             \\ 
   TUM-GAID~\cite{Hofmann2014TheTG}                & 15$\times$         & 14$\times$               & 1           & $\checkmark$          &             &             &             \\ 
   OU-LP~\cite{Iwama_IFS2012}                   & 200$\times$        & 33$\times$             & 2           &           &             &             &             \\ 
   OU-MVLP~\cite{takemura2018multi}                 & 515$\times$        & 1114$\times$           & 14          &           &             &             &             \\ 
   \textbf{VersatileGait} & \textbf{550$\times$} & \textbf{4300$\times$} & \textbf{33} & \textbf{$\checkmark$} & \textbf{$\checkmark$}   & \textbf{$\checkmark$}   & \textbf{$\checkmark$}   \\ 
   \bottomrule
   \label{datasetcompare}
   \end{tabular}
   \end{table}

\subsection{Synthetic Dataset}
Synthetic datasets have been widely used in various tasks in computer vision for three main advantages as follows:

\paragraph{Explore More Applications} 3D model is utilized to generate synthetic data for accurate human depth estimation and human part segmentation~\cite{Varol_2017_CVPR}. A virtual dataset~\cite{fabbri2018learning} is collected to tackle with multi-person tracking and  pose estimation under occlusion condition.

\paragraph{Mitigate the Lack of Data} A segmentation dataset~\cite{Hu_2019_CVPR} is proposed to mitigate the lack of occlusion-aware annotation. Likewise, a synthetic logo dataset~\cite{Su2017DeepLL} is presented to enrich the real manual labeled dataset.

\paragraph{Enhance the Performance on Real Dataset} This fashion such as~\cite{8954024,9102822, 8237485, Zhao2019MultisourceDA,zheng2020structured3d}, employs supervised learning and domain adaptation to achieve superior results. With the help of~\cite{Wang_2019_CVPR}, better performance of crowd counting is achieved. Additionally, pretraining with synthetic data~\cite{Tremblay2018TrainingDN} improves the performance of models.

%% file: data/data.tex
\section{The VersatileGait Dataset}

The goal of our VersatileGait is to introduce a large-scale gait dataset with fine-grained attributes and complicated scenarios to satisfy various research demands and practical applications. This section presents the key challenges of synthetic data collection and details of our data generation pipeline. Moreover, we summarize the properties of our dataset for further studies.

\subsection{Key Challenges} Generating high-quality synthetic data with fine-grained attributes and complicated scenarios raises many challenges: 
1) \textit{How to represent a walking individual?}
Generally, a walking individual is represented as appearance (human with accessories) and action (walking style).  We bind a 3D human model and walking animations to simulate the combination of appearance and action. 
2) \textit{How to select the attributes of individuals?} Since silhouettes mainly represent the contours of pedestrians, we select several contour-sensitive attributes such as genders and walking styles. 
3) \textit{How to simulate the complicated scenario?} 
We define the complicated scenario as the viewpoints (horizontal and vertical) of cameras and the number of pedestrians in the scene. Then, we simulate that with a game engine.
4) \textit{How to guarantee the quality of silhouettes?} 
The data processing, such as background subtraction and segmentation, is inevitable in the process of
real silhouette collection, which causes poor image quality.
To tackle this problem, we use a dark background and white 3D human models without texture to directly synthesize silhouettes without any processing. 
Specifically, we illustrate our data generation pipeline as follows. 

\subsection{Dataset Generation}
\label{collection}
\begin{figure}[t]
\centering
\vspace{-0.8cm}  
        \setlength{\abovecaptionskip}{-0.01cm}   
        \setlength{\belowcaptionskip}{-0.2cm}

\includegraphics[width=8.5cm]{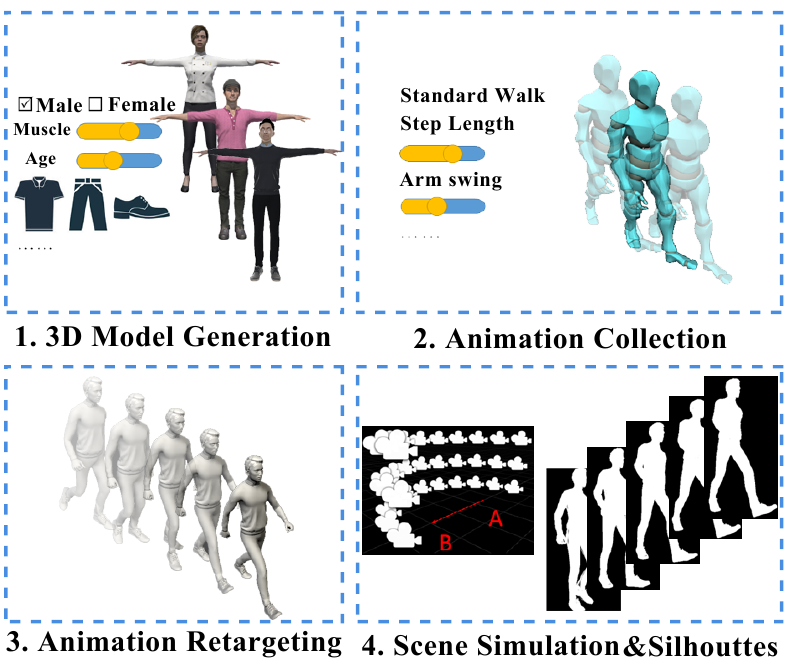}
\caption{The data generation pipeline of VersatileGait}

\label{fig:pipe}
\end{figure}

Our pipeline for generating synthetic data consists of four steps, as illustrated in \cref{fig:pipe}.

\noindent\textit{\textbf{1) 3D Model Generation}} To get more realistic human models, we use Make Human~\cite{bastioni2008ideas} for the model generation. It is a frequently-used open-source tool for modeling parameterized 3D characters, which are highly correlated with the attribute parameters (\eg, genders, ages, heights) and strictly restricted by human morphology to make the human body more realistic. 
With the randomly generated parameters on Make Human, we create numbers of models in a unified pose for easy manipulation. Besides, to reflect attributes distinction on silhouettes, we add some artificial constraints, \eg, people of different genders have different preferences for accessories.
Finally, we generate 150 realistic 3D human models with balanced attribute distribution to represent the appearance of walking individuals.

\noindent\textit{\textbf{2) Walking Animation Collection}} Mixamo~\cite{mixamo} is an online platform containing numerous built-in skeletal animations, which are commonly used to animate 3D characters.
We collect some human walking animations from the platform to simulate the walking style of mankind, such as standard walking, brutal walking.
Besides, to increase the diversity of our dataset, we adjust the step stride and arm angle of the animations to represent different walking habits.
Altogether, we collect 100 human walking animations, which are also taken as a special attribute related to the temporal information of an individual's action.

\noindent\textit{\textbf{3) Animation Retargeting }} Animation retargeting aims at binding animations to a targeted human model, which is essential for the realistic and smooth gait sequence. 
We use the Mecanim animation system in Unity3D~\cite{unity} to establish a connection between the models' structure and the default humanoid bones. Then, we manually correct the binding errors for the sake of fluent and natural movements, resulting in 11,000 high-quality walking individuals.

\noindent\textit{\textbf{4) Scene Simulation \&  Silhouettes Capture}} To satisfy the practical demands of complicated scenarios and get high-quality silhouettes, we design scenarios in Unity3D, which is widely used in game development. 
Specifically, we use dark skyboxes as the background and adopt 6 orthogonal parallel lights for projection. There are up to 3 models in the scene at the same time.
These textureless human models walk between specified starting and ending points under 33 cameras with 11 horizontal views, and 3 vertical views (\ie, pitch angles). 
These cameras directly capture binary silhouettes without any extra processing to ensure the high quality of our VersatileGait, and the annotations are automatically generated through the pipeline.
As a result, we generate 72 million frames, with a resolution of 280 $\times$ 200, grouped into one million synthetic gait sequences with the aforementioned 11,000 subjects. 

\label{generation}

\subsection{Properties of VersatileGait}
\label{properties}
As a synthetic dataset, VersatileGait has five remarkable properties compared to existing datasets.

\begin{figure}[htbp]
\centering
        \setlength{\abovecaptionskip}{-0.01cm}   
        \setlength{\belowcaptionskip}{-0.2cm}
        \subfigure[The comparison of synthetic data~\cite{wang2020surpassing} and real data~\cite{zheng2015scalable} in person ReID. There is a significant difference between them, due to the texture, hue, \etc.]{
        \hspace{-5mm}
        \includegraphics[width=7.5cm]{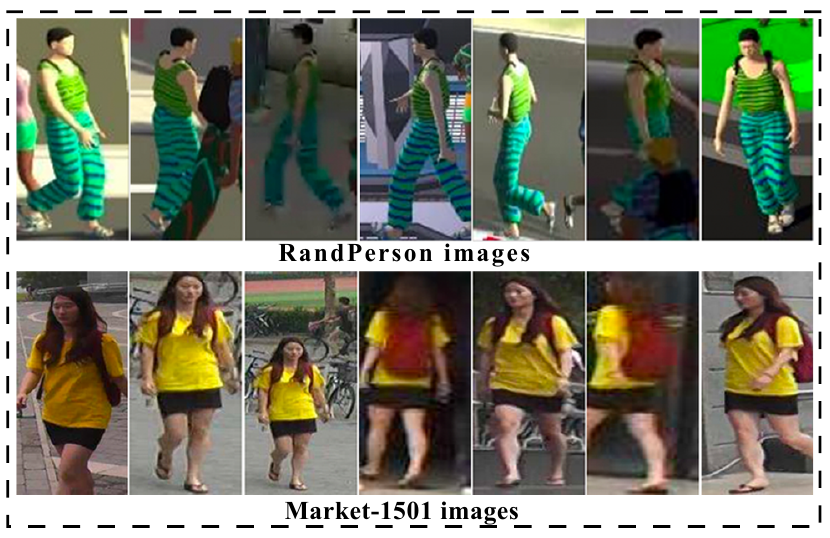}
        }
        \quad
        \subfigure[The comparison of synthetic data and real data~\cite{takemura2018multi,1699873} in gait recognition. Synthetic data are more realistic and high-quality thanks to textureless and colorless properties. It demonstrates the advantages of using synthetic data in the field of gait recognition.]{
        \hspace{-5mm}
        \includegraphics[width=7.5cm]{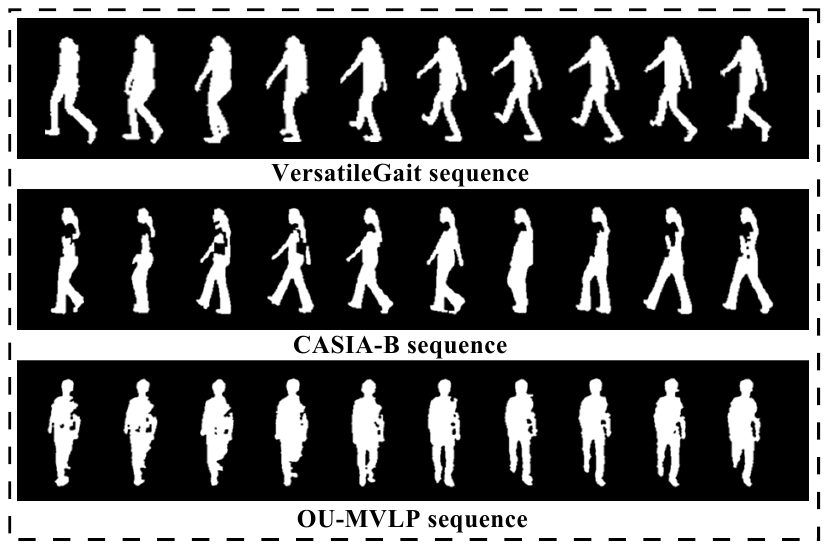}
        }
        \quad
        \caption{The comparison of synthetic and real datasets in gait recognition and other fields.}
        \label{domaingap}
\end{figure}

\noindent\textit{\textbf{1) Small Domain Gap}} When it comes to synthetic datasets, it is inevitable to confront the issue of domain gap. As shown in \cref{domaingap}, the severe domain gap exists in person ReID datasets due to the color and texture. By contrast, since silhouettes contain no color or texture information but contours of individuals, employing synthetic datasets with silhouettes introduces less domain gap.

\noindent\textit{\textbf{2) Fine-Grained Annotations}} VersatileGait contains fine-grained annotations such as genders and walking styles, which could be used for diverse applications.

\noindent\textit{\textbf{3) High Quality}}  The synthetic data generation pipeline directly produces silhouettes without any data processing such as background subtraction or segmentation, which would cause severe image quality damage.

\noindent\textit{\textbf{4) Complicated Scenarios}} As far as we know, VersatileGait is the first public dataset that contains the multi-pitch angles and the multi-person gait scenario, which is very common in practical gait recognition problem but absent in existing real datasets.

\noindent\textit{\textbf{5) Large Amount}} We use the proposed pipeline to synthesize a dataset consisting of 11000 subjects with more than one million sequences, which is the largest gait dataset containing the change of viewpoints. Besides, we release our data generation toolkit and users could utilize this to synthesize customized data to satisfy their research demands.

%% file: data/effective.tex
\section{Dataset Effectiveness Studies}
Before introducing our VersatileGait to various applications, we conduct several experiments to validate the effectiveness of our datasets in the following steps: 1) we explore the effects of different dataset size; 2) we design several situations to evaluate the effects under the different mix ratio of real data and the synthetic data; 3) we analyze the effects of two-stage training strategies.

\begin{figure}[htb]
\centering
\includegraphics[width=8cm]{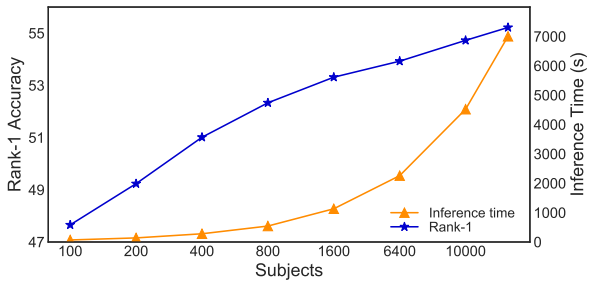}
\caption{Averaged rank-1 accuracies and inference time cost under different amounts of the training dataset. Note that the scale of gallery subjects will also increase while the training dataset gets larger.}

\label{fig:data}
\end{figure}

\paragraph{The Effects of Dataset Size}  Larger datasets provide more diverse knowledge. However, as a retrieval task, the inference time of gait recognition increases dramatically as the size of the dataset increases. Therefore, we conduct several experiments to quantitatively explore the performance gain and inference time cost with respect to the size of the dataset. Specifically, we pretrain GaitSet~\cite{chao2019gaitset} model on VersatileGait with size from 100 to 10000 and evaluate the model on CAISIA-B~\cite{1699873} without finetuning, and we keep the division between the training set and the test set unchanged. As shown in Fig~\ref{fig:data},  as the size of the dataset becomes larger, inference time increases dramatically while the performance gain becomes smaller. Thus, by utilizing this finding, researchers could select the proper dataset size, according to the research demands and hardware platform.

\paragraph{The Effects of the Mix Training} Mix training~\cite{Nowruzi2019HowMR} with real data and synthetic data is an efficient way to improve the performance of deep learning models. To explore the effects of mix training, we design experiment settings as follows: first, we keep the training dataset size fixed and use five different mix ratios of real and synthetic data. Specifically, the real data ratio varies between 2.5\%, 5\%, 10\%, 50\%, 100\%. Then the rest data is filled up with VersatileGait. Based on these experiment settings, we train GaitSet~\cite{chao2019gaitset} models to show the performance gains from VersatileGait. The more detailed experiment settings are shown in the supplementary material. As shown in \cref{ratio}, it indicates two phenomena: 1) VersatileGait considerably compensates the performance decay when real data scarcity is severe; 2) we achieve comparable performance to the model trained with all real data, by only using 50\% real data filled up with the synthetic data.


\begin{figure}[t]
    \centering
    \hspace{-1cm}
    \setlength{\abovecaptionskip}{-0.01cm}   
    \setlength{\belowcaptionskip}{-0.2cm}   
    \includegraphics[width=8cm]{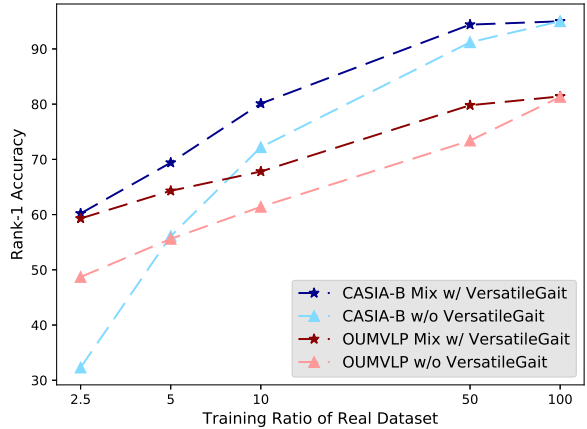}
    \caption{Results for mix training with VersatileGait and real dataset. The performance is tested on corresponding real test dataset. The horizontal axis is in a log scale.}
\label{ratio}
\end{figure}

\begin{table*}[]
   \caption{Averaged rank-1 accuracies on \textbf{CASIA-B} under different conditions, excluding identical-view cases. The conditions include origin, pretrain model with VersatileGait, mix the whole real training set with VersatileGait for training.}
   \centering
\small
   \begin{tabular}{|c|l|l|l|l|l|l|l|l|l|l|l|l|l|l|}
   \hline
   \multicolumn{3}{|c|}{Gallery NM\#1-4}                               & \multicolumn{11}{c|}{0$^{\circ}$-  180$^{\circ}$}                                                                                                                                               & \multicolumn{1}{c|}{\multirow{2}{*}{mean}} \\ \cline{1-14}
   Method                    & Probe                    & Condition    & 0$^{\circ}$            & 18$^{\circ}$           & 36$^{\circ}$           & 54$^{\circ}$           & 72$^{\circ}$           & 90$^{\circ}$           & 108$^{\circ}$          & 126$^{\circ}$          & 144$^{\circ}$          & 162$^{\circ}$          & 180$^{\circ}$          & \multicolumn{1}{c|}{}                      \\ \hline\hline
   \multirow{9}{*}{GaitSet}           & \multirow{3}{*}{NM\#5-6} & origin                     & 90.8          & 97.9          & \textbf{99.4} & \textbf{96.9} & 93.6          & 91.7          & 95.0          & 97.8          & 98.9          & 96.8          & 85.8          & 95.0                                       \\ \cline{3-15} 
                                                 &                          & pre+finetune               & 92.8          & 97.9          & \textbf{99.4} & 98.6          & \textbf{95.2} & \textbf{93.1} & 96.2          & \textbf{98.8} & 98.7          & 97.9          & 89.5          & 96.2                                       \\ \cline{3-15} 
                                                 &                          & mix+finetune               & \textbf{93.7} & \textbf{98.6} & 99.0          & 98.4          & 95.1          & 92.9          & \textbf{96.7} & 98.7          & \textbf{99.0} & \textbf{98.2} & \textbf{90.0} & \textbf{96.4}                              \\ \cline{2-15}
                                                 & \multirow{3}{*}{BG\#1-2} & origin                     & 83.8          & 91.2          & 91.8          & 88.8          & 83.3          & \textbf{81.0} & 84.1          & 90.0          & 92.2          & \textbf{94.4} & 79.0          & 87.2                                       \\ \cline{3-15} 
                                                 &                          & pre+finetune               & 87.1          & 91.4          & 92.8          & 90.5          & \textbf{86.0} & 80.0          & 84.7          & \textbf{90.4} & 92.5          & 92.0          & 80.9          & 88.0                                       \\ \cline{3-15} 
                                                 &                          & mix+finetune               & \textbf{87.7} & \textbf{91.9} & \textbf{93.3} & \textbf{90.6} & 84.2          & 79.0          & \textbf{84.9} & 92.0          & \textbf{95.4} & 92.2          & \textbf{81.6} & \textbf{88.5}                              \\ \cline{2-15}
                                                 & \multirow{3}{*}{CL\#1-2} & origin                     & 61.4          & \textbf{75.4} & \textbf{80.7} & \textbf{77.3} & \textbf{72.1} & \textbf{70.1} & \textbf{71.5} & 73.5          & 73.5          & 68.4          & 50.0          & 70.4                                       \\ \cline{3-15} 
                                                 &                          & pre+finetune               & \textbf{64.0} & 72.7          & 76.7          & 75.0          & 68.4          & 68.9          & 71.1          & \textbf{73.6} & \textbf{76.3} & \textbf{80.4} & \textbf{60.0} & \textbf{70.6}                              \\ \cline{3-15} 
                                                 &                          & mix+finetune                & 62.1          & 74.3          & 78.0          & 77.2          & 69.7          & 67.7          & 70.4          & 73.4          & 74.3          & 76.9          & 59.2          & \textbf{70.6}                                       \\ \hline\hline
   \multirow{9}{*}{GaitPart}                     & \multirow{3}{*}{NM\#5-6} & origin                     & 94.1          & 98.6          & 99.3          & 98.5          & 94.0          & 92.3          & 95.9          & 98.4          & 99.2          & 97.8          & 90.4          & 96.2                                       \\ \cline{3-15} 
                                                 &                          & pre+finetune               & 95.5          & 98.4          & 99.6          & 98.6          & 96.0          & 92.4          & 96.1          & 98.5          & 99.3          & 98.1          & 92.7          & 96.8                                       \\ \cline{3-15} 
                                                 &                          & mix+finetune               & \textbf{95.8} & \textbf{98.7} & \textbf{99.8} & \textbf{98.7} & \textbf{96.3} & \textbf{92.6} & \textbf{96.3} & \textbf{98.6} & \textbf{99.6} & \textbf{98.5} & \textbf{92.9} & \textbf{97.1}                              \\ \cline{2-15} 
                                                 & \multirow{3}{*}{BG\#1-2} & origin                     & 89.1          & 94.8          & 96.7          & \textbf{95.1} & \textbf{88.3} & \textbf{94.9} & 89.0          & 93.5          & 96.1          & 93.8          & 85.8          & 91.5                                       \\ \cline{3-15} 
                                                 &                          & pre+finetune               & 91.2          & 95.4          & 95.0          & 94.0          & 87.6          & 83.9          & 89.1          & \textbf{94.3} & 95.7          & 94.2          & 86.9          & 91.6                                       \\ \cline{3-15} 
                                                 &                          & mix+finetune               & \textbf{91.5} & \textbf{95.8} & \textbf{95.1} & 94.0          & 87.4          & 84.3          & \textbf{89.3} & 94.0          & \textbf{96.2} & \textbf{94.3} & \textbf{87.1} & \textbf{91.8}                              \\ \cline{2-15} 
                                                 & \multirow{3}{*}{CL\#1-2} & origin                     & 70.7          & \textbf{85.5} & \textbf{86.9} & \textbf{83.3} & 77.1          & \textbf{72.5} & 76.9          & \textbf{88.2} & 83.8          & 80.2          & 66.5          & 78.7                                       \\ \cline{3-15} 
                                                 &                          & pre+finetune               & \textbf{73.5} & 85.1          & 85.3          & 81.4          & \textbf{78.3} & 70.6          & 77.0          & 81.6          & 84.2          & \textbf{81.5} & \textbf{68.5} & \textbf{78.9}                              \\ \cline{3-15} 
                                                 &                          & mix+finetune               & 73.3          & 85.0          & 85.6          & 81.5          & 78.1          & 71.1          & \textbf{77.1} & 81.4          & \textbf{84.4} & 81.2          & 68.4          & \textbf{78.9}                              \\ \hline
\end{tabular}
\label{premix}
\end{table*}

\paragraph{The Effects of two-stage Training} Generally, the performance of the model will be improved considerably if pretrained with a dataset with diversity. To evaluate the diversity of VersatileGait, we adopt two two-stage training strategies to advance the performance of existing powerful methods~\cite{chao2019gaitset,Fan_2020_CVPR}: 1) VersatileGait for pretraining; 2) mix the whole real train dataset with VersatileGait for pretraining. Both of them are followed with finetuning operation and tested on the corresponding real test dataset. As shown in \cref{premix}, it indicates the performance of mainstream methods will increase by a considerable margin, thus VersatileGait is a practical dataset for pretraining. Further, we find it effective to help real-world gait recognition using VersatileGait from three aspects mentioned above.
\label{DA}

%% file: data/applications.tex
\section{Diverse Applications of VersatileGait}
Based on VersatileGait, we explore practical applications in two aspects. Firstly, we utilize fine-grained attributes to conduct attribute-guided gait recognition in \cref{attr}, which makes the gait recognition system more accurate and the retrieval (the inference stage of gait model) faster. Secondly, we focus on the scenario under multi-pitch angles in \cref{multi-pitch}, where we explore the importance of the multi-pitch angle scenario and the relation between each pitch angle.

\subsection{Attribute-Based Applications}
\label{attr}

\paragraph{Attribute Selection Criterion}

The silhouettes contain no texture and color and are only related to the contour of the subjects. Therefore, only attributes corresponding to the contour can be further used in gait recognition. Followed by this criterion, genders and walking styles are used for further exploration. We utilize these fine-grained attributes to improve gait recognition from two perspectives: accuracy and speed.

\begin{figure}[t]
   \centering
   \includegraphics[width=8.1cm]{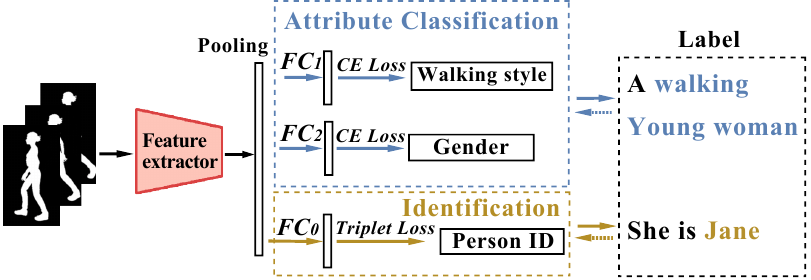}
   \caption{The framework of attribute guided gait recognition, which follows the multi-task learning~\cite{Ruder2017AnOO} architecture. These two attribute classifiers are composed of only two fully connected layers and a ReLU activation.}
   
   \label{fig:loss}
   \end{figure}

\paragraph{More Accurate Gait Recognition} 
The fine-grained attribute labels could play the role of auxiliary supervision for gait recognition.  To simplify the experiment, we use the network of GaitSet~\cite{chao2019gaitset} but remove the multi-scale partition module as our baseline method. Then we expand the baseline method with simple attribute classifiers resulting in a multi-task learning framework, which is shown in~\cref{fig:loss}. For convenience, we let $A$ be the set of attributes and let $L_{a}$ represent the cross-entropy loss of different attributes. Then, our optimization objective can be formulated as~\cref{lossf}.

\begin{equation}
   \textit{L}_{total} = \textit{L}_{triplet} + \sum_{a \in A}{\lambda_{a} L_{a}}
\label{lossf}
\end{equation}

The gender and walking style can be discriminated well by our method, which achieves the accuracies of 88.8\% and 99.63\% for attribute prediction, respectively. We report the detailed results of gait recognition in~\cref{ml}, which shows that the performance of our baseline model is improved by a large margin under the multi-task learning framework. It shows the selected attributes have a significant promotion effect on the gait recognition task by introducing the supervision of fine-grained labels.

   \begin{table*}[t]
      \caption{Rank-1 accuracy of the baseline method and the attribute-guided method tested on the CASIA-B.}
      
      \begin{center}
      \begin{tabular}{|c|l|l|l|l|l|l|l|l|l|l|l|l|l|}
      \hline
      \multicolumn{2}{|c|}{Gallery NM\#1-4}       & \multicolumn{11}{c|}{0°-  180°}                                                                                                                                               & \multicolumn{1}{c|}{\multirow{2}{*}{Mean}} \\ \cline{1-13}
      \multicolumn{1}{|l|}{Probe} & Method        & 0°            & 18°           & 36°           & 54°           & 72°           & 90°           & 108°          & 126°          & 144°          & 162°          & 180°          & \multicolumn{1}{c|}{}                      \\ \hline
      \hline
      \multirow{2}{*}{NM\#5-6} & Baseline                & 46.7          & 46.8          & 54.4          & 59.1          & 59.8          & 57.7          & 55.0          & 55.2          & 58.5          & 56.5          & 45.8          & 54.1                                       \\ \cline{2-14} 
                               & Attribute-guided          & \textbf{80.6} & \textbf{79.2} & \textbf{87.0} & \textbf{90.2} & \textbf{88.4} & \textbf{84.8} & \textbf{81.6} & \textbf{83.9} & \textbf{88.4} & \textbf{89.8} & \textbf{80.6} & \textbf{84.9}                              \\ \hline\hline
      \multirow{2}{*}{BG\#01}  & Baseline                & 46.2          & 45.7          & 52.7          & 56.2          & 57.1          & 52.4          & 50.6          & 51.2          & 55.2          & 53.7          & 44.7          & 51.4                                       \\ \cline{2-14} 
                               & Attribute-guided           & \textbf{78.2} & \textbf{77.7} & \textbf{85.4} & \textbf{89.1} & \textbf{87.0} & \textbf{81.8} & \textbf{79.2} & \textbf{81.8} & \textbf{86.4} & \textbf{88.0} & \textbf{79.2} & \textbf{83.1}                                       \\ \hline\hline
      \multirow{2}{*}{CL\#01}  & Baseline                & 31.3          & 31.3          & 38.0          & 42.1          & 43.4          & 39.5          & 39.1          & 38.9          & 40.3          & 39.8          & 31.5          & 37.7                                       \\ \cline{2-14} 
                               & Attribute-guided           & \textbf{57.3} & \textbf{57.0} & \textbf{66.3} & \textbf{73.1} & \textbf{73.7} & \textbf{69.9} & \textbf{68.3} & \textbf{68.4} & \textbf{70.6} & \textbf{68.7} & \textbf{59.0} & \textbf{66.6}                              \\ \hline
      \end{tabular}
      \end{center}
      \label{ml}
\end{table*}

\paragraph{Faster Retrieval} Gait recognition needs to calculate the similarity between the query and the numerous gallery instances in the inference phase. It is bothered by the heavy computation cost, which slows down the speed of inference. To solve this problem, we propose to reduce the search space using the results of attributes prediction. As we have stated above, attribute prediction gets high accuracies for gender and walking style. Therefore, we use these two attributes as the criterion to reduce the search space. For simplicity, we use a single attribute for reduction each time.

We treat the selected attribute of the instance as reliable if its confidence score is higher than the threshold. If the attribute of the query instance is reliable, we will use the gallery instances, which have the same reliable attribute and its value as the query instance, resulting in the search space reduction. For example, given a query instance's walking style (brutal walking) is reliable, we choose all the gallery instances that have the same reliable attribute of walking style (brutal walking) for matching. Otherwise, the corresponding search space will include all the gallery instances.

\begin{figure}[htbp]
   \centering
   \includegraphics[width=8cm]{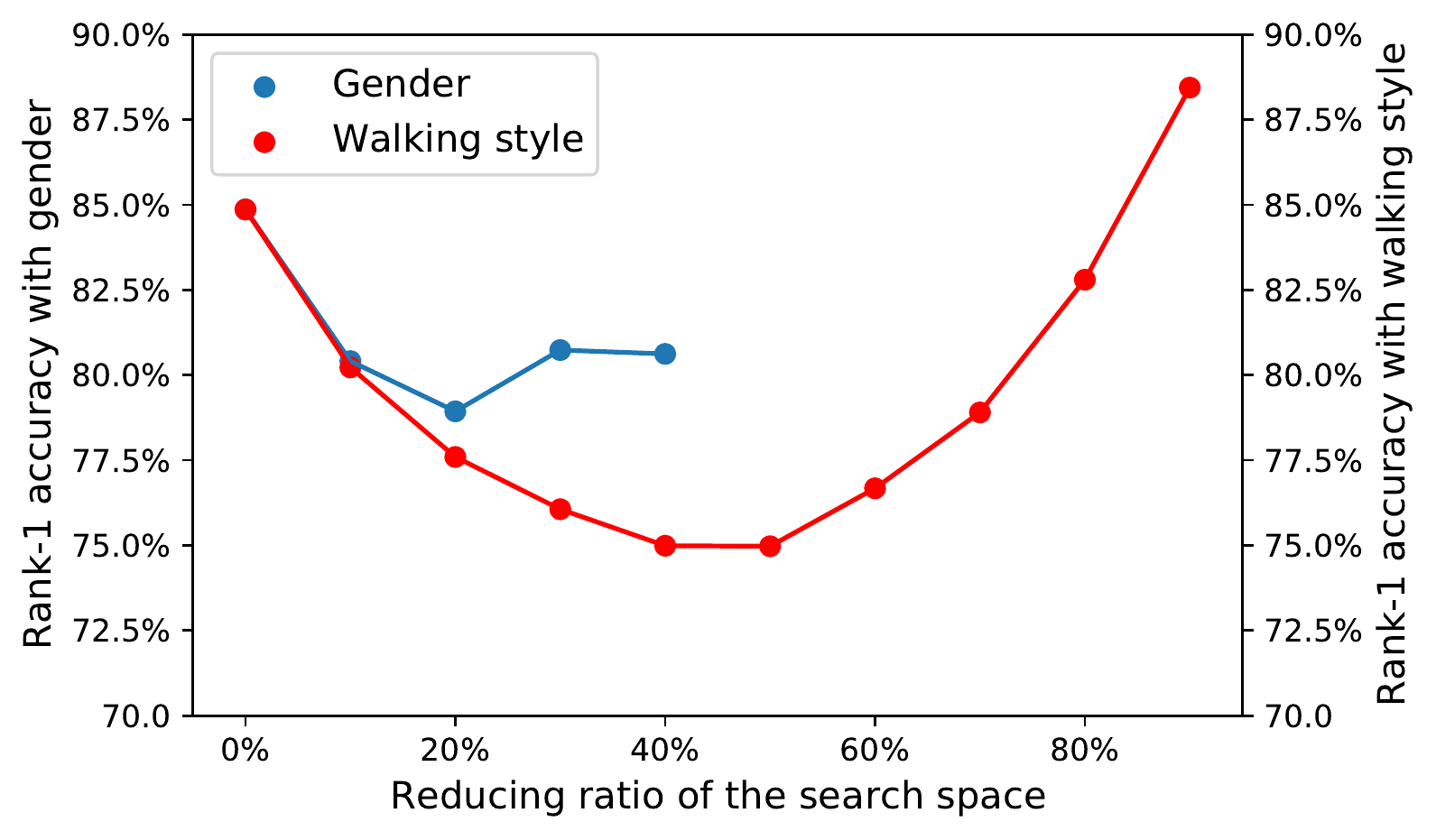}
   \caption{Rank-1 accuracy on VersatileGait when using different attributes to accelerate the retrieval.}
   
   \label{fig:retrieval}
   \end{figure}

The curves of the relationship between the search space scale and the rank-1 accuracy could be seen in~\cref{fig:retrieval}. With this strategy, we could speed up the gait recognition by more than two times while the accuracies decrease by no more than 10\% and 4\%, guided by walking styles and genders, respectively. It shows that: 1) using attributes as the search restriction, we could reduce the scale of gallery subjects with a tolerable performance drop of accuracy; 2) there is no definite relationship between accuracy and the scale of reduced search space. Sometimes, the accuracy could even be improved.

\subsection{Multi-Pitch Angle Gait Recognition}
\label{multi-pitch}
\begin{figure}[t]
   \centering
   \includegraphics[width=8cm]{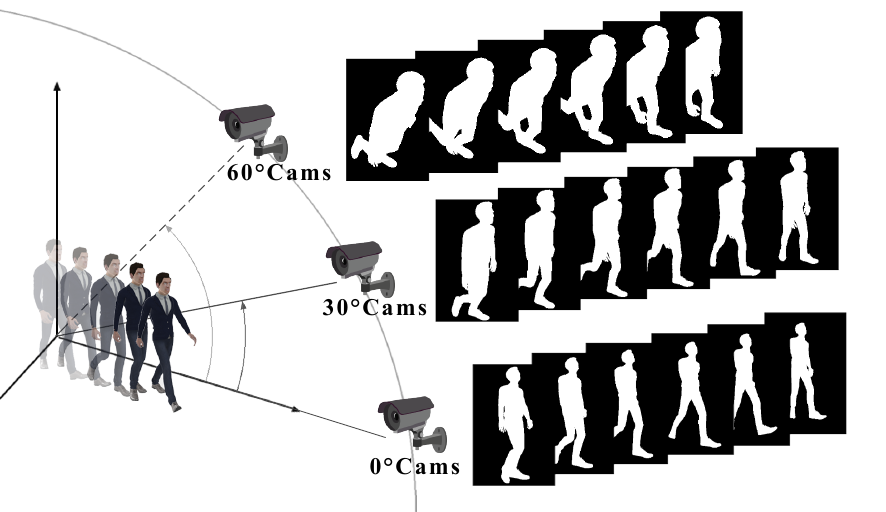}
   \caption{Multi-pitch angle scenario. Three sequences in the cameras of different pitch angles, where pedestrians are distorted significantly.}
   \label{fig:multiheight}
   \end{figure}
Multi-pitch angles of cameras can cause severe distortion in real gait recognition scenarios but are not considered among the previous gait dataset. To analyze the impact of this kind of distortion, we evaluate the performance of GaitSet on our VersatileGait for the cross-pitch angle recognition problem.

\paragraph{The Necessity of Multi-Pitch Angle Data} Existing datasets ignore the multi-pitch angle scenarios, which may cause severe geometric distortion on silhouettes. To figure out the necessity of multi-pitch angle data, we conduct a pair of comparative experiments using the existing methods. 

\begin{figure}[t]
   \hspace{-1cm}
   \centering
   \includegraphics[width=7.5cm]{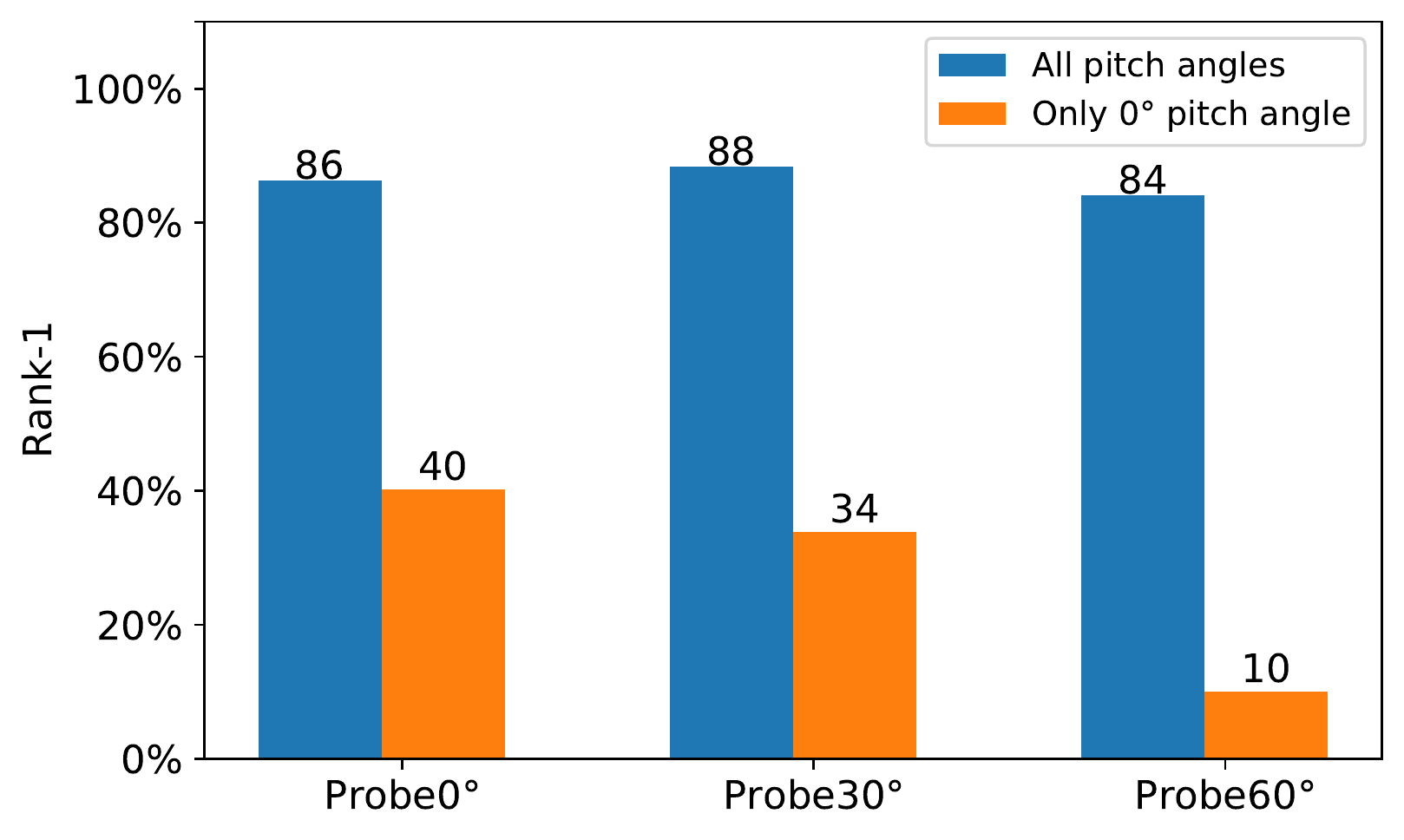}
   \caption{The results of GaitSet tested on multi-pitch angle data.}
   \label{0andall}
   \end{figure}

As shown in \cref{0andall}, the averaged cross-pitch angle performance of GaitSet~\cite{chao2019gaitset} will drop significantly if the multi-pitch angle data is not provided for training. 
Therefore, the collection of multi-pitch angle data is essential to gait recognition in the wild.

\begin{figure}[t]
   \centering
   \includegraphics[width=8cm]{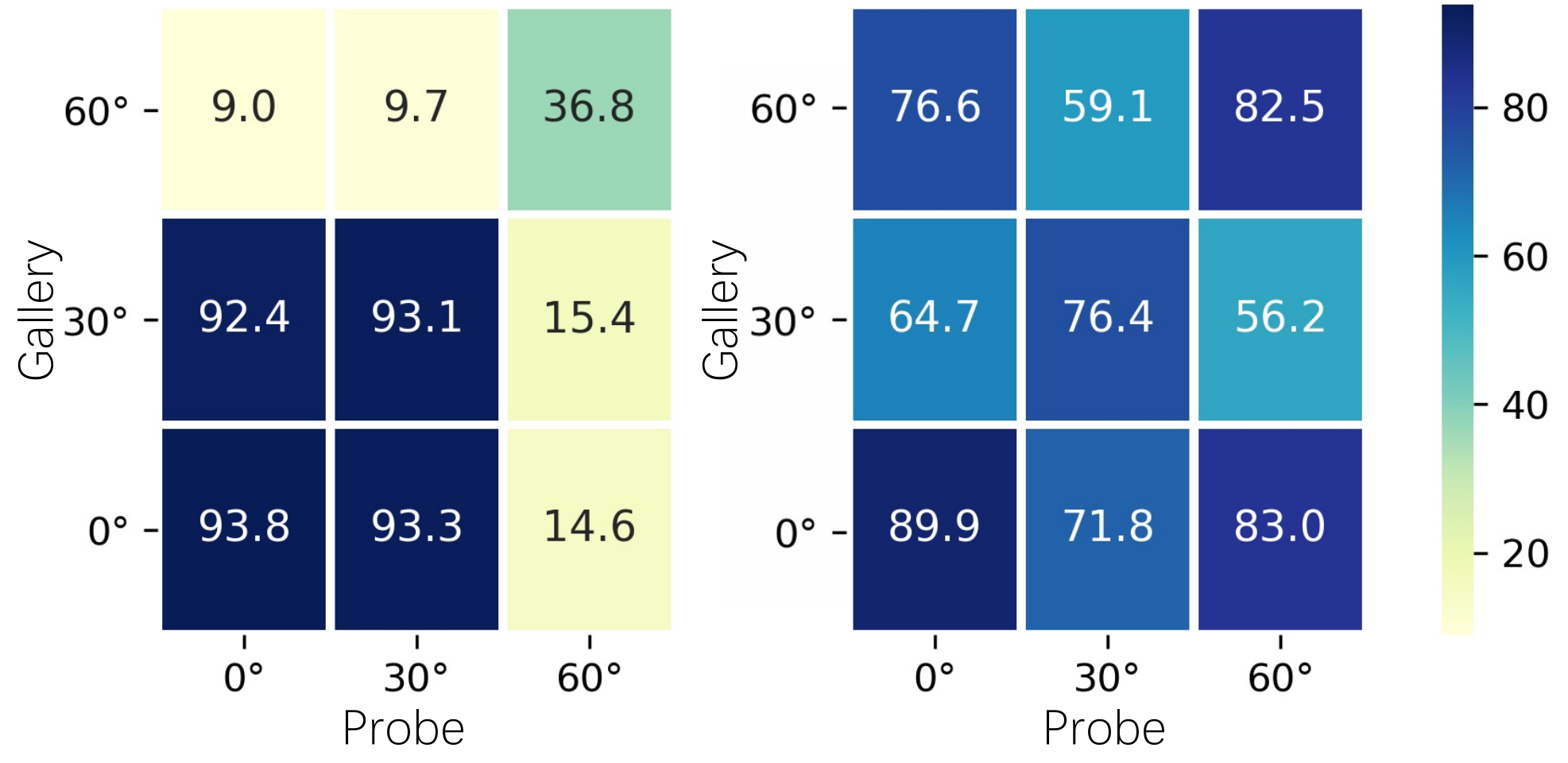}
   \caption{The rank-1 accuracies of GaitSet for the cross-pitch angle recognition on VersatileGait. \textbf{Left:} train model without 60$^{\circ}$ data; \textbf{Right:} train model without 30$^{\circ}$ data.}
   \label{lackdata}
   \end{figure}

\label{multi-pitch}
\paragraph{The Effects of Different Pitch Angle} The results above show models are not robust enough to multi-pitch angle scenario. However, the different pitch angles may play a different role in gait recognition. Therefore, we conduct two experiments to explore the effect of lacking different pitch angle data: 1) train the model with the data of pitch angles 0$^{\circ}$, 30$^{\circ}$, excluding 60$^{\circ}$; 2) train the model with the data of pitch angles 0$^{\circ}$ and 60$^{\circ}$ data, excluding 30$^{\circ}$. 
We test the rank-1 accuracies of GaitSet for the cross-pitch angle and cross-view recognition, then the dimension of the horizontal view is averaged for visualization. As shown in \cref{lackdata}, if the training set lacks higher pitch angle data, the performance of the model degrades dramatically. By contrast, if the training set lacks relatively low pitch angle data, the model maintains a relatively stable performance.

\begin{figure}[t]
   \centering
   \includegraphics[width=8.2cm]{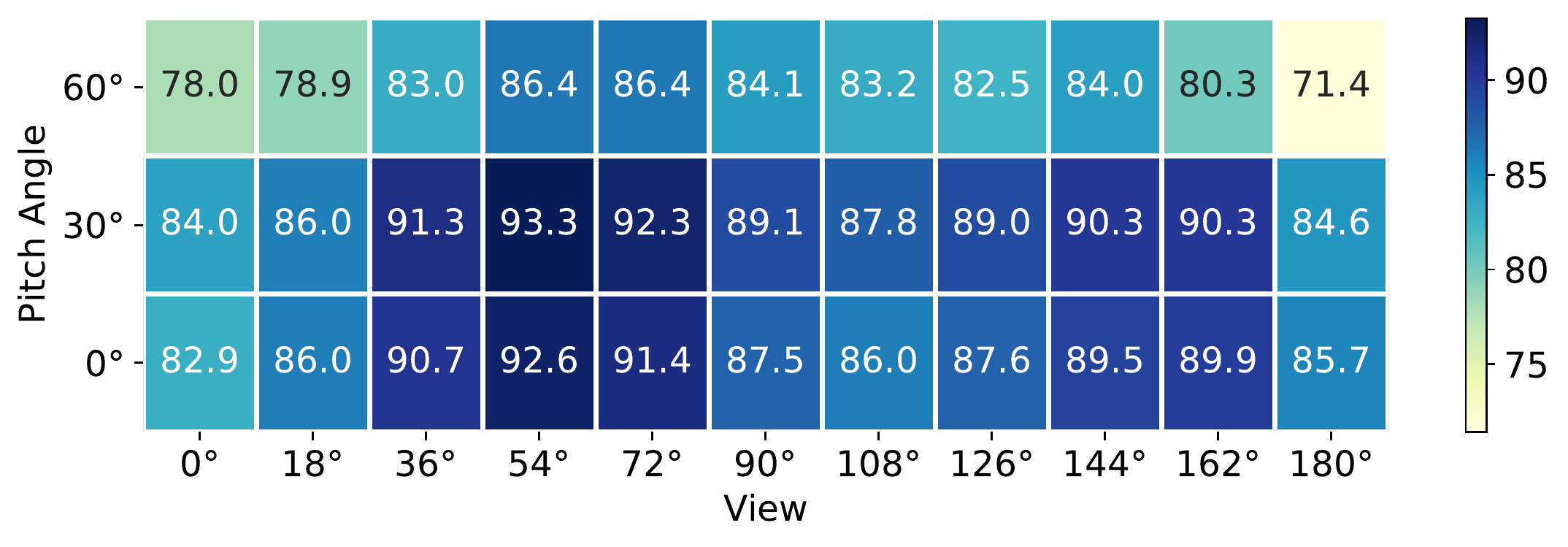}
   \caption{The averaged rank-1 accuracies of GaitSet on VersatileGait for the 33 kinds of probes from different viewpoints.}
   \label{comprehensive}
   \end{figure}

   \paragraph{The Analysis of Cross Viewpoints Gait Recognition} In practical scenarios, there are cameras with various viewpoints. 
   Therefore, we take both the 11 horizontal views and 3 pitch angles together into consideration.
   Concretely, we test the performance of GaitSet on VersatileGait for all possible viewpoint pairs (33 probe viewpoints vs. 33 gallery viewpoints). As shown in~\cref{comprehensive}, the recognition accuracy decreases significantly on the probe of 60$^{\circ}$ pitch angle even the training data is provided. Besides, the cross horizontal view problem is severer when the cross pitch angle factor is included, especially on the view around 0$^{\circ}$ and 180$^{\circ}$.

%% file: data/future.tex
\section{Potential Applications and Dissusion}

\begin{figure}[t]
  \centering
  \includegraphics[width=7.5cm]{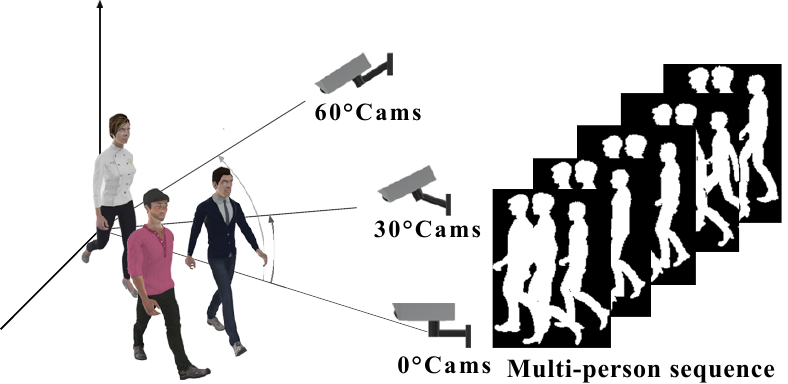}
  \caption{Multi-person gait. The silhouettes are severely occluded, which brings new challenges on gait recognition.}
  \label{multiperson}
  \end{figure}

  \paragraph{Multi-Person Gait Recognition}
  Existing public datasets mainly focus on a single pedestrian scenario. However, multi-person walking cases are common in the real world, which is shown in \cref{multiperson}. To cover this scenario, we simulate the scenarios with the person up to three at the same time. We will release this multi-person gait dataset for further exploration. This application scenario may point to new directions for future research. 

\paragraph{Disentangled Representation Learning} 


 We can combine the same subject with different variables (e.g., accessories and walking styles). With the properties of our dataset, there are two perspectives of disentanglement. 1) Conducting disentanglement learning following the previous methods~\cite{Li_2020_CVPR,gait-recognition-via-disentangled-representation-learning} that decompose a subject into the gait-related feature and accessories-related feature. We can control the change of carriers and generate numerous training pairs; 2) A subject can be further disentangled into the individual-related feature and walking-style related feature. Based on this disentanglement paradigm, we can research the characteristics of the more intrinsic feature. More details can be seen in the supplementary material.


%% file: data/conclusion.tex
\section{Conclusion}

In this paper, we have proposed a high-quality large-scale synthetic gait dataset named VersatileGait rendered by a game engine,
which contains much more fine-grained attributes and complicated scenarios than those of existing gait
datasets. VersatileGait is composed of around one million silhouette  sequences  of  11,000  subjects.
Based on VersatileGait, we have conducted a variety of learning effectiveness studies to improve the mainstream methods by a considerable margin.
Besides, we enrich various applications including attribute guided gait recognition with multi-task learning and gait retrieval 
acceleration by fast attribute filtering. Moreover, we have evaluated gait recognition performance in the new scenario of multi-pitch angles. Extensive experiments have shown the great potential of  
VersatileGait in both the research community and the industry.